\documentclass[conference]{IEEEtran}
\IEEEoverridecommandlockouts

\usepackage{graphicx}

\usepackage{cite}
\usepackage{amsmath,amssymb,amsfonts}

\usepackage{graphicx}
\usepackage{textcomp}
\usepackage{xcolor}
\usepackage{cite}
\usepackage{amsmath,amssymb,amsfonts}

\usepackage{graphicx}
\usepackage{textcomp}
\usepackage{xcolor}
\usepackage{cite}
\usepackage{float}
\usepackage{hyperref}
\usepackage{amsmath,amssymb,amsfonts}
\usepackage{graphicx}
\usepackage{textcomp}
\usepackage{xcolor}
\usepackage{scalerel}
\usepackage{tikz}

\usepackage{algcompatible}
\usepackage{algpseudocode}
\usepackage{algorithm}
\usetikzlibrary{svg.path}

\definecolor{orcidlogocol}{HTML}{A6CE39}
\tikzset{
  orcidlogo/.pic={
    \fill[orcidlogocol] svg{M256,128c0,70.7-57.3,128-128,128C57.3,256,0,198.7,0,128C0,57.3,57.3,0,128,0C198.7,0,256,57.3,256,128z};
    \fill[white] svg{M86.3,186.2H70.9V79.1h15.4v48.4V186.2z}
                 svg{M108.9,79.1h41.6c39.6,0,57,28.3,57,53.6c0,27.5-21.5,53.6-56.8,53.6h-41.8V79.1z M124.3,172.4h24.5c34.9,0,42.9-26.5,42.9-39.7c0-21.5-13.7-39.7-43.7-39.7h-23.7V172.4z}
                 svg{M88.7,56.8c0,5.5-4.5,10.1-10.1,10.1c-5.6,0-10.1-4.6-10.1-10.1c0-5.6,4.5-10.1,10.1-10.1C84.2,46.7,88.7,51.3,88.7,56.8z};
  }
}

\newcommand\orcidicon[1]{\href{https://orcid.org/#1}{\mbox{\scalerel*{
\begin{tikzpicture}[yscale=-1,transform shape]
\pic{orcidlogo};
\end{tikzpicture}
}{|}}}}
\def\BibTeX{{\rm B\kern-.05em{\sc i\kern-.025em b}\kern-.08em
    T\kern-.1667em\lower.7ex\hbox{E}\kern-.125emX}}
\usepackage[utf8]{inputenc}
\def\BibTeX{{\rm B\kern-.05em{\sc i\kern-.025em b}\kern-.08em
    T\kern-.1667em\lower.7ex\hbox{E}\kern-.125emX}}
\begin{document}

\title{ProdRev: A DNN framework for empowering customers using generative pre-trained transformers\\}

\author{
\IEEEauthorblockN{Aakash Gupta}
\IEEEauthorblockA{\textit{Think Evolve Consultancy LLP} \\
\textit{Navi Mumbai, India}\\
aakash@thinkevolveconsulting.com}\\
\and 
\IEEEauthorblockN{Nataraj Das\orcidicon{0000-0003-0796-6521}}
\IEEEauthorblockA{\textit{Dept. of Computer Science and Engineering} \\
\textit{National Institute Of Technology}\\
Agartala, India \\
natarajdas9@gmail.com}

}

\maketitle

\begin{abstract}
Following the pandemic, customers, preference for using e-commerce has accelerated. Since much information is available in multiple reviews (sometimes running in thousands) for a single product, it can create decision paralysis for the buyer. This scenario disempowers the consumer, who cannot be expected to go over so many reviews since its time consuming and can confuse them. Various commercial tools are available, that use a scoring mechanism to arrive at an adjusted score. It can alert the user to potential review manipulations. This paper proposes a framework that fine-tunes a generative pre-trained transformer to understand these reviews better. Furthermore, using "common-sense" to make better decisions. These models have more than 13 billion parameters. To fine-tune the model for our requirement, we use the curie engine from generative pre-trained transformer (GPT3). By using generative models, we are introducing abstractive summarization. Instead of using a simple extractive method of summarizing the reviews. This brings out the true relationship between the reviews and not simply copy-paste. This introduces an element of "common sense" for the user and helps them to quickly make the right decisions. The user is provided the pros and cons of the processed reviews. Thus the user/customer can take their own decisions.
\end{abstract}

\section{Introduction}
The paper illustrates a methodology for empowering customers through a deliberate decision-making process through fine-tuning a generative pre-trained transformer. There has been a global shift of users towards e-commerce platforms. This trend was accelerated during the covid-19 induced lock-downs. These platforms also provide extensive information for the concerned product in user reviews. By analyzing them, users can anticipate the quality of the product. Reviews provide essential details regarding the product and its performance in real life. However, the user can easily get overwhelmed with too much information. Such an overload can cause decision paralysis and lead to users making sub-optimal decisions. We propose an architecture based on deep learning techniques to mitigate this scenario and empower the customers.
\begin{itemize}
    \item We fine-tune a GPT3 model from OpenAI.
    \item This architecture helps the user to get a clear understanding of the product reviews by summarizing and clustering the reviews into pros and cons.
    \item The generative transformer model also provides a final verdict about the concerned product.
    \item Our proposed framework aims at minimizing misleading machine verdicts. By providing transparent summarization of the reviews as pros and cons, the user can make their own decisions.
\end{itemize}
For ease of understanding, we have structured the paper into the following sections. In Section II we provide a literature review on the research conducted in this space. Section III provides a broad overview of our working methodology, which includes data set preparation. In section IV we discuss the results of our experiments and section V provides a summary and future scope for research.
\begin{figure*}[t!]
    \centering
    \includegraphics[width=0.74\textwidth,keepaspectratio]{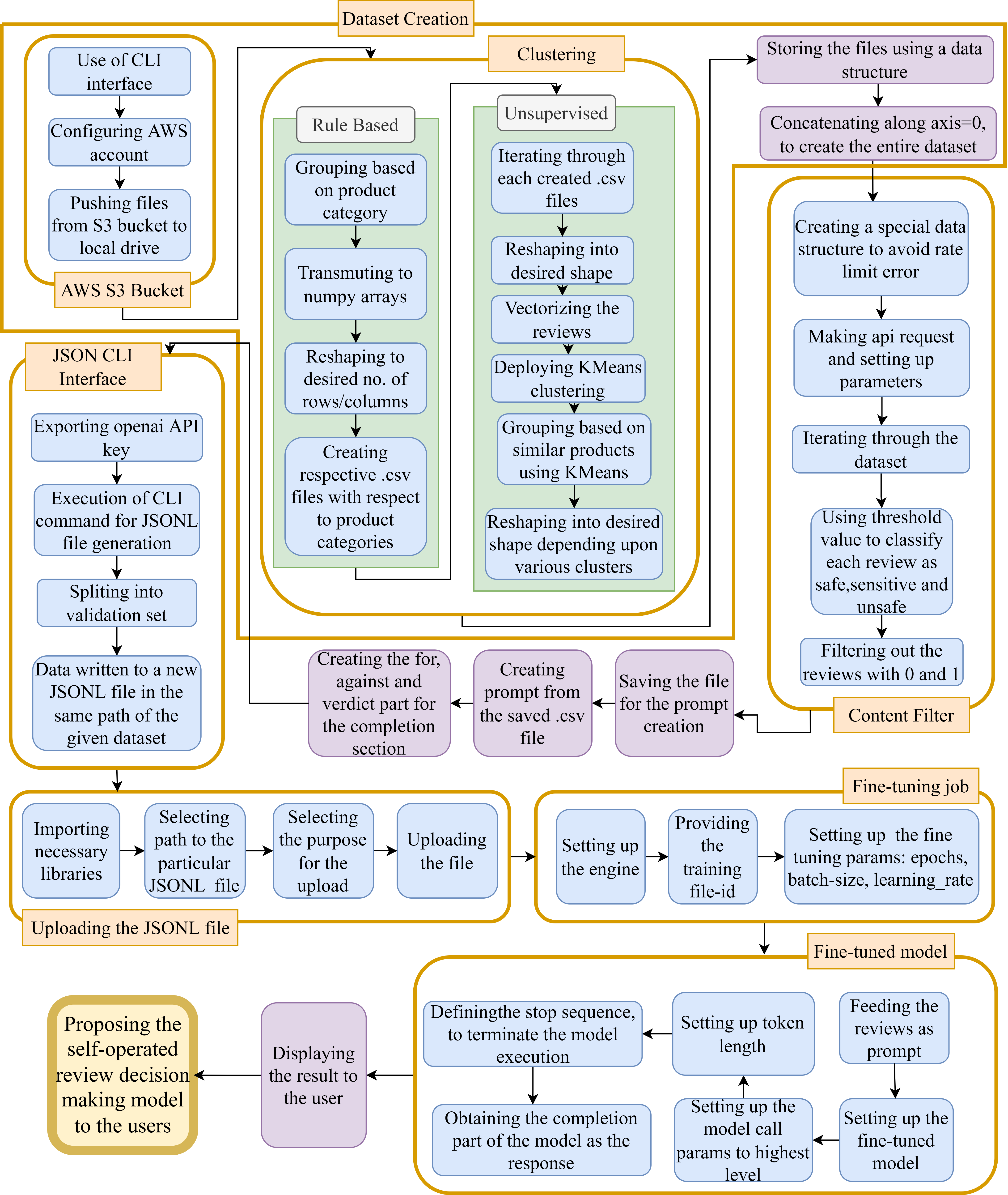}
    \caption{An overview of the working framework}
    \label{overview}
\end{figure*}
\section{Literature review}
\cite{alt-etal-2019-fine} introduces generative pre-trained transformer(GPT). These models can capture semantic and syntactic features through an attention selective mechanism. By extending the standard transformer architecture with an attentive selection mechanism for the multi-instance learning scenario -- we can achieve state-of-art performance on predicting distinct relationships with high confidence.\\
In \cite{gptnature} we runs three tests, to check how well GPT3 performs with logico-mathematical, semantic, and ethical requests. GPT3, is a third-generation language model, that is auto-regressive. This transformer uses deep learning techniques, that are capable of producing texts, at par with humans. The paper also discusses the various limits and consequences of using the pre-trained transformer.\\
The language model presented in this paper\cite{brown2020language}, has been trained on 175 billion parameters, which performs outstandingly well on various natural language processing tasks and various other benchmarks. The paper also introduces the concept of zero-shot, one-shot, and few-shot learning, which gives an example to the model about what completion part needs to create and what should be the structure of the output. It also provides various ways to manipulate the output, produced by the language model. The paper also discusses the various ways, where the model can be turned into a zero-shot  model, for a specific purpose, so that one can achieve better performance.\\

 \cite{10.1007/978-3-030-90321-3_61} talks about the art of fine-tuning the GPT3 model, for a particular task, that outperforms other state-of-art in that field. Here the fine-tuning job was summarising Russian text, which was tested upon BERT-based metrics.
\section{Approach}
Humans tend to use short-term memory for reviewing information and making a decision on that basis. It often turns out to be a cumbersome task for the human brain to review something heterogeneous in nature, especially when the item to be reviewed is large. Such situations lead to poor decision-making, not only in terms of conclusion but also the time taken to come up with one. Here comes the product of our research, which makes the review items homogeneous and provides a final verdict depending upon the pros and cons, thus assisting the human decision-making process. Making the reviews \cite{FELDMANNWUSTEFELD2014108}, tend to make better neural connections in the human brain, providing the user, the opportunity to analyze the pros and cons more efficiently, and thus making the user independent in terms of final decision making. As the model also provides a final verdict depending upon its understanding of the reviews, one can also compare how well he or she is using his/her decision making abilities.
\subsection{Synopsis of the working methodology}
The figure \ref{overview} depicts the data flow of the stated methodology.
We start with data preparation. Since the dataset can contain unsafe words, we need to filter them out. The next step is prompt creation. The reviews are structured such that the model performs better during the fine-tuning process. The OpenAI fine-tuning endpoint accepts the training data in a specific file format. The dataset is then converted from CSV file format to a jsonl file format. Once converted, the jsonl file is uploaded to the fine-tuning API endpoint. This process generates a file-id that can track the fine-tuning process. We optimize the model by experimenting with different hyper-parameters \ref{table}. We also train with multiple (train) dataset sizes \ref{rouge} \ref{bert}. The final generative fine-tuned transformer is then available for deployment. 

\subsection{Data set preparation}
The data set chosen\cite{Amazon} was of reviews about various products that were fetched from the AWS S3 bucket, after configuring the concerned AWS account. After pushing all the S3 bucket contents from the cloud to the local drive, based on product category, the files were sorted out.
\begin{algorithm}
\caption{Preparation of data set}\label{dataset}
\begin{algorithmic}[1]
\State Storing review files in terms of product category.
\State \textbf{path\_1}: {The path to local drive}
\For {Files in \textbf{path\_1}:}
\If {$review\_len \ge 120:$}
 \State ${extract, toarray(), reshape, tocsv(), Save}$
\EndIf
\EndFor
\State \textbf{path\_2}:{The path to saved reshaped files}
\State Initialising instance of \textbf{KMeans} cluster 
\State $ k= 90 $
\For {File in \textbf{path\_2}:}
\State Empty List declaration
\State  toarray(), reshape, vectorize
\State  \textbf{KMeans}: fit\_transform()
\For {cluster in clusters(k=90):}
\State Reshaping
\State Appending the list
\EndFor
\State Concatenating the list
\State Saving the File to desired location
\EndFor
\State \textbf{path\_3}:{The path to saved clustered files}
\State Empty list declaration
\For {files in path\_3:}
\State Appending files to the list 
\EndFor
\State Concatenating the list
\State Saving the file as the final data set.
\Return {X}
\end{algorithmic}
\end{algorithm}
The algorithm \ref{dataset} depicts a complete algorithmic approach towards the creation of the data set. Here it can be seen to manage such a large data set, unsupervised technique, KMeans clustering\cite{9666257} has been deployed to cluster\cite{5453745} reviews of the same kind, such that reviews of different products don't get mixed up. Finally, the files of different clustered product categories were concatenated to create the final data set.\\
The data set created consists of 70963 rows, each having 15 reviews. Each row in the data set can be treated as 15 reviews for a single product.
\subsection{Filtering the contents in the data set}
Since the reviews were taken from the public domain, it is necessary to see, if the reviews were safe in terms of ethical, political, and religious sentiments. So for this content filtration, it is necessary to pass each review through a filtration function, as provided by Open Ai, that labels each review as 0,1,2 corresponding to safe, sensitive, and unsafe content. It is mandatory to remove out the ones with label 2. However discarding the reviews with label 2 has got a particular threshold value\cite{9641704} of -0.355, which is nothing but a threshold value imposed upon the natural logarithm of the probability of the label is 2.
\begin{equation}
    logprob_{x}= log_{e} P(x)
\end{equation}
\begin{equation}
    thresh= -0.355
\end{equation}
\begin{equation}
    Action =
    \begin{bmatrix}
     Reject\, ,logprob_{2} \ge thresh\\
    Hold\, ,logprob_{2} < thresh\\
    \end{bmatrix}
    \label{eq3}
\end{equation}
Thus equation \ref{eq3} has been followed to discard the unsafe reviews. By the term hold, it has been depicted that for such situations, the log probs of labels 0 and  1 were taken into account,  and the final label is provided depending on the higher value of log prob as compared between 0 and 1.
\subsection{Designing the prompt section}
\begin{figure}[t!]
    \centering
    \includegraphics[width=0.53\textwidth, keepaspectratio]{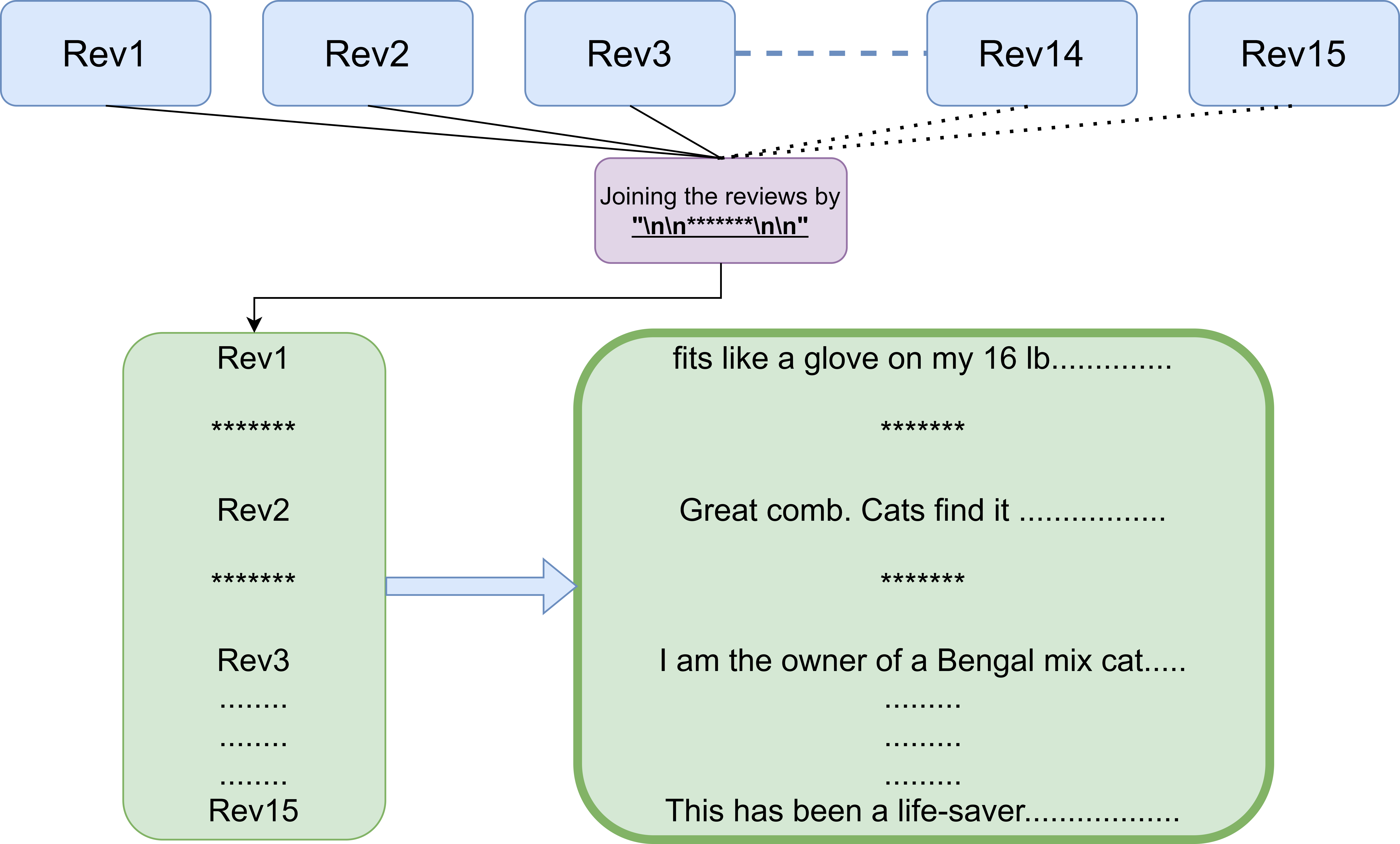}
    \caption{Prompt Creation}
    \label{prompt}
\end{figure}
The above figure \ref{prompt} shows the structure of the prompt that has been used for the fine-tuning purpose. In the original dataset, the 15 reviews were joined through the string \textbf{”\textbackslash n\textbackslash n*******\textbackslash n\textbackslash n”}, such that the desired structure for the prompt gets created. The structure of the prompt is not hard-coded, it can change depending upon the principles of prompt engineering.
\subsection{Designing of the completion section}
The completion section creation was rather a difficult task that has been done manually through the human annotators. There was a manual annotation of each review, corresponding to the structure of the completion section. The deployment of the human workforce made the annotations comprehensive and generative, rather than an extractive summary.Such actions contributed towards human-level machine performance, i.e. the completion part generated by the model resembles human-level performance and accuracy. The completion part consists of 3 parts: \textbf{pros}, \textbf{cons} and \textbf{verdict}. Making these 3 sections available to the user, makes it comprehensive, and helps the user to make an informed decision, within a  short time. The user can also confirm his or her final decision from the verdict section produced by the model. At the end, as stop sequence was inserted in order to stop further execution of the model as well as to manipulate the generated completion part

\begin{table}[H]
\caption{Hyper parameters for training the transformer}
\centering
\begin{tabular}{||c c ||} 
 \hline
Param & Value \\ [0.5ex] 
 \hline\hline
Engine & Curie \\ 
 \hline
 batch\_size & 49 \\
 \hline
n\_epochs & 5 \\
 \hline
learning\_rate & 0.1 \\
 \hline
 use\_padding & True \\ 
 \hline
\end{tabular}
\vspace{1ex}

\label{table}
\end{table}

\subsection{Modelling and transmitting the  JSONL file}
Once the prompt and completion .csv file is ready, the next in line is to convert the csv files to jsnol file. For fine tuning purpose Open AI allows only .jsonl files. So using CLI commands the above prepared data set gets converted to the desired file format.
\begin{figure}[t!]
    \centering
    \includegraphics[width=0.49\textwidth,keepaspectratio]{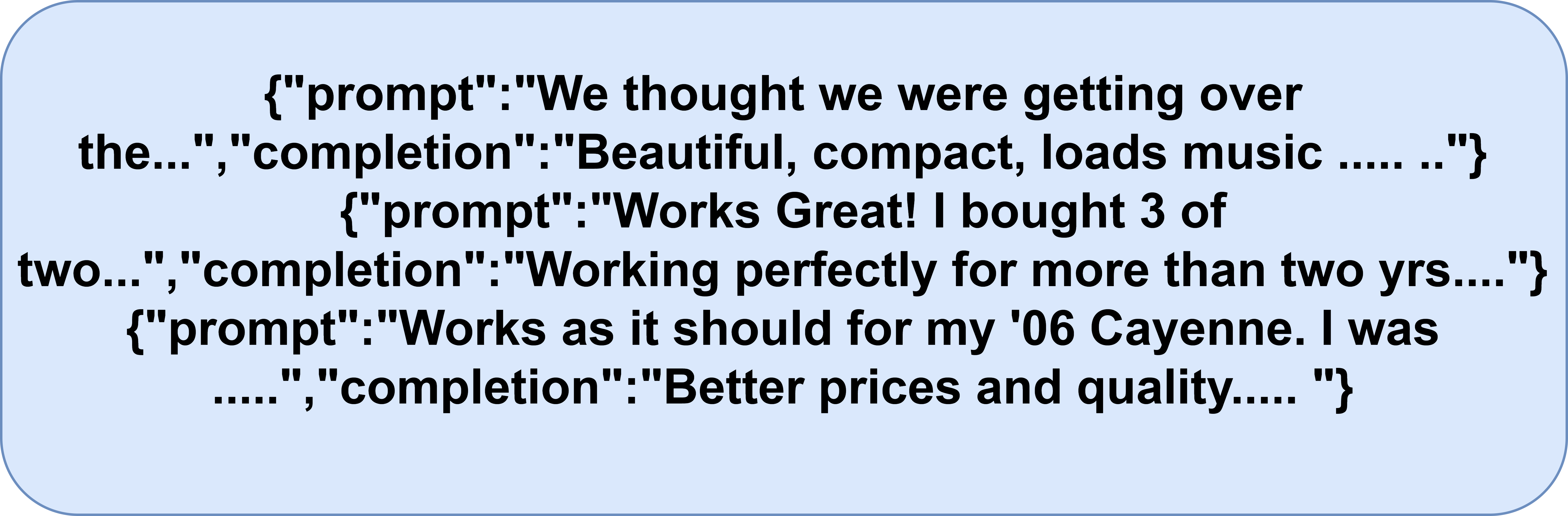}
    \caption{JSONL file format}
    \label{json}
\end{figure}
The figure\ref{json} depicts the jsonl file format that has been used for the fine-tuning purpose. Once ready, the file was uploaded to the Open AI cloud, with the purpose of fine-tuning.
\subsection{Accomplishing the fine-tuning job}
Once the JSON file was uploaded, next in line is the final fine-tune job\cite{LEE2020101983}. The generative pre-trained transformer is now trained on a particular data set for a very particular purpose. This step is the final leap where the transformer, learns about what sort of output it needs to generate when fed with a particular prompt. The hyperparameters for training the transformer were selected as such \ref{table}.

\section{Viable result analysis}
In order to analyse the performance of the model, it has been put through a lot of tests and training, in order to make it more robust and better at producing desirable outputs.
\subsection{Deployment in real life usage}
The final trained model that has been fine-tuned on 485 data points, was deployed in a real life use case, where there was heterogeneous information about a particular product, and the model was meant to generate a homogeneous output along with a final verdict.
\begin{figure}[t!]
    \centering
    \includegraphics[width=0.45\textwidth,keepaspectratio]{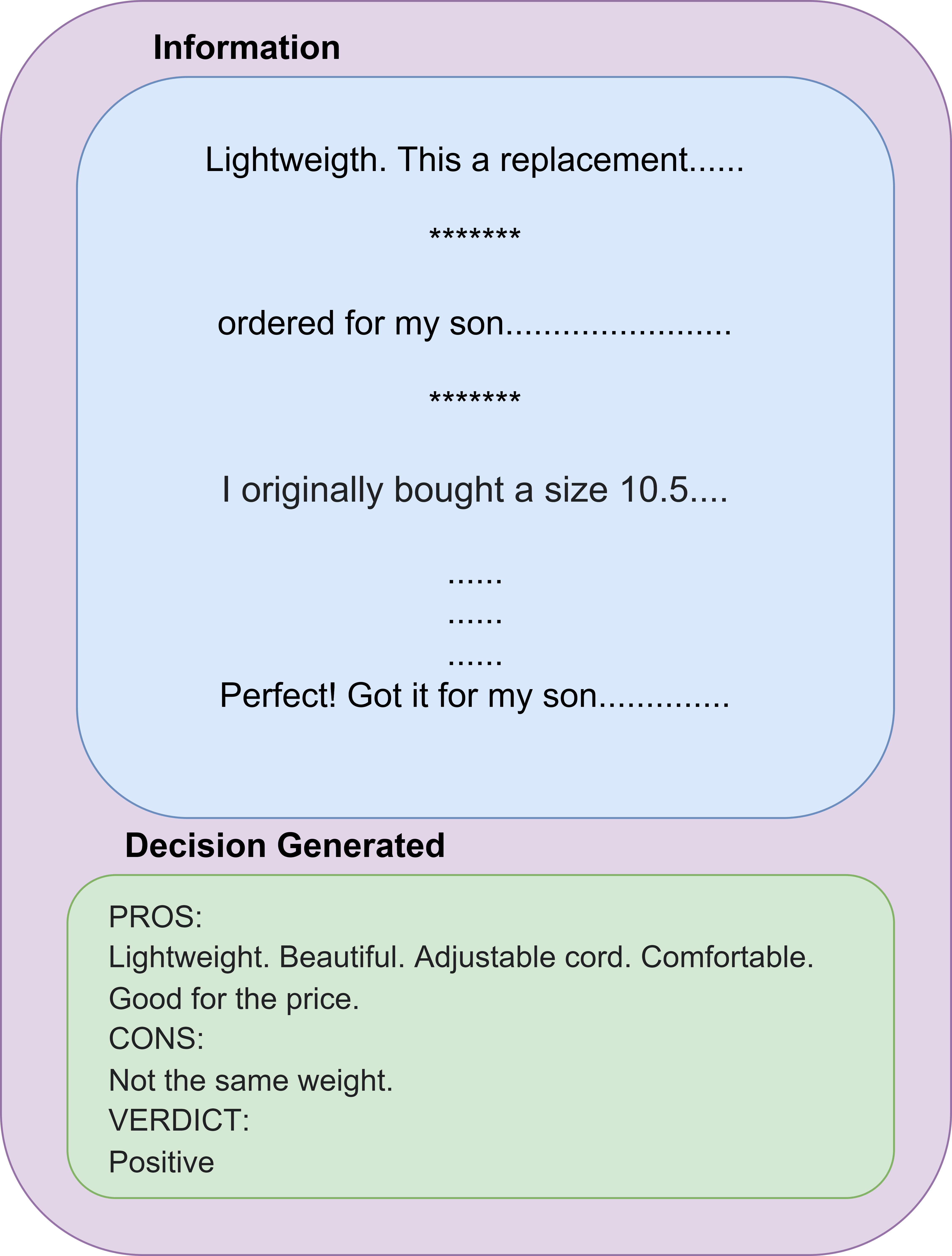}
    \caption{Inference on Out-of-sample reviews}
    \label{real}
\end{figure}

The figure \ref{real} clearly depicts how well the model performs in generating a homogeneous completion section, which is not only precise, but highly comprehensive in nature. Such output allows a user to view directly the nitty-gritty of a product. The model also produces a final verdict\cite{senti}, depending upon its own understanding about the pros and cons, thus providing great assistance to the user in decision making.

\subsection{Bert\_Score Analysis}
\begin{figure}[t!]
    \centering
    \includegraphics[width=0.41\textwidth,keepaspectratio]{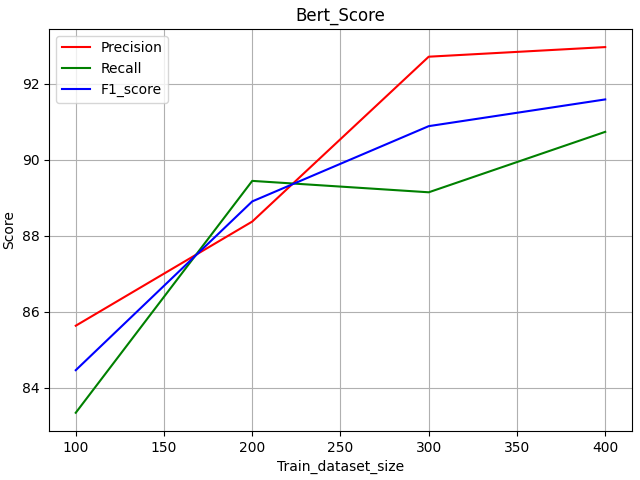}
    \caption{Bert\_Score Analysis}
    \label{bert}
\end{figure}

Further analysis of the training procedure included calculation of performance metric, in order to understand, a trend in the training process vs the train data size. In order to do so, Bert\_Score\cite{bert-score} was calculated, as it deploys similarity of the tokens through contextual embedding. Following the trend, the final model was trained upon 485 data set, that performs pretty well. The calculation of Bert\_Score involved various train data size, in order to estimate, how much of training, can produce desirable results. However, from the figure\ref{bert} it can be clearly seen that even with small data set size, the fine-tuned model performed pretty well.
\subsection{Rouge\_Score Analysis}
\begin{figure}[t!]
    \centering
    \includegraphics[width=0.41\textwidth, keepaspectratio]{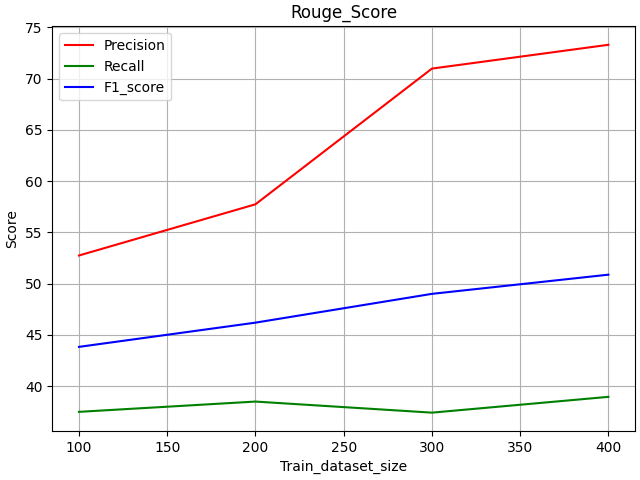}
    \caption{Rouge\_Score Analysis}
    \label{rouge}
\end{figure}

Rouge score analysis\cite{lin-2004-rouge} was also carried out in order to support the results produced by the bert\_score analysis as in figure\ref{rouge}. Considering the demand of the proposed methodology, unigram rouge score was considered, in order to estimate a trend in the training process.
The decrease in rouge score can be understood from the fact that rougue uses unigram matching of words rather than the context or sense of the generated output. However it confirms the trend that has been followed for training and fine-tuning the transformer. 
\section{Conclusion}
The methodology presented in the paper is highly robust and efficient in comprehending and producing a homogeneous completion section, that becomes easy for human brains to percept. The model also produces quite an accurate verdict, depending upon its understanding of the pros and cons, thus providing high assistance towards decision making. The model, which is a pre-trained transformer has been fine-tuned on datasets that have been manually annotated by, the human workforce. Such a step allowed the model to understand the comprehensive nature of the completion section, that further contributed towards the generation of output, which is at par with human capability.\\
However, such manual annotation of the dataset is highly time-consuming. This was the very reason for fine-tuning the final model with only 485 data points. Throughout the stated methodology, availability of the desired dataset was a big hurdle to deal with.\\
Even with such a small data set, the transformer was able to achieve high performance and produce robust, precise, and aesthetically pleasing results, that allow the user to quickly arrive at a decision, just by looking at the produced pros and cons against the particular entity. Future scope for this research work included training it on a larger data set, which will make the model more accurate. Experimentation with other pre-trained architecture framework as well as hyperparameters tuning can provide better results.
\section*{Acknowledgement}
The authors would like to acknowledge the Mae Fah Luang University, for their immense support towards the success of this research content. The university not only provided a platform to publish the research findings, but also gave financial assistance towards publication of this paper in terms of registration waive off .
\bibliographystyle{unsrt}
\bibliography{reference}
\end{document}